\pdfoutput=1

\documentclass[11pt]{article}

\usepackage[]{ACL2023}
\usepackage{makecell}
\usepackage{times}
\usepackage{latexsym}
\usepackage{tcolorbox}
\usepackage{multirow}
\usepackage[T1]{fontenc}
\usepackage{footmisc}

\usepackage[utf8]{inputenc}

\usepackage{microtype}

\usepackage{inconsolata}
\usepackage{verbatim}
\usepackage{booktabs}
\usepackage{amsmath}
\usepackage{floatrow}
\usepackage{longtable}
\usepackage{booktabs}
\newfloatcommand{capbtabbox}{table}[][\FBwidth]
\title{\raisebox{-0.1cm}{\includegraphics[width=0.5cm]{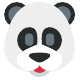}} PANDA: Preference Adaptation for Enhancing\\ Domain-Specific Abilities of LLMs}

\author{
  An Liu\textsuperscript{1}, Zonghan Yang\textsuperscript{1}, Zhenhe Zhang\textsuperscript{1}, Qingyuan Hu\textsuperscript{1}, Peng Li\textsuperscript{*,2},\\
  \textbf{Ming Yan\textsuperscript{3}, Ji Zhang\textsuperscript{3}, Fei Huang\textsuperscript{3}, Yang Liu\textsuperscript{*,1,2,4}} \\
  $^1$Dept. of Comp. Sci. \& Tech., Institute for AI, Tsinghua University, Beijing, China \\
  $^2$Institute for AI Industry Research (AIR), Tsinghua University, Beijing, China \\
  $^3$Institute of Intelligent Computing, Alibaba Group \\
  $^4$Jiangsu Collaborative Innovation Center for Language Competence, Jiangsu, China \\
  \texttt{la22@mails.tsinghua.edu.cn; lipeng@air.tsinghua.edu.cn}\\
  \texttt{liuyang2011@tsinghua.edu.cn}}

\begin{document}
\maketitle

\DefineFNsymbols*{1}{*}
\setfnsymbol{1}

\renewcommand{\thefootnote}{\fnsymbol{footnote}} 
    \footnotetext[1]{Corresponding authors: Peng Li and Yang Liu.}
\renewcommand{\thefootnote}{\arabic{footnote}}

\begin{abstract}
While Large language models (LLMs) have demonstrated considerable capabilities across various natural language tasks, they often fall short of the performance achieved by domain-specific state-of-the-art models. One potential approach to enhance domain-specific capabilities of LLMs  involves fine-tuning them using corresponding datasets. However, this method can be both resource and time-intensive, and not applicable to closed-source commercial LLMs.
In this paper, we propose \textbf{P}reference \textbf{A}daptation for E\textbf{n}hancing \textbf{D}omain-specific \textbf{A}bilities of LLMs (PANDA), a method designed to augment the domain-specific capabilities of LLMs by leveraging insights from the response preference of expert models without requiring fine-tuning.
Our experimental results reveal that PANDA significantly enhances the domain-specific ability of LLMs on text classification and interactive decision tasks. Moreover, LLM with PANDA even outperforms the expert model that being learned on 4 tasks of ScienceWorld. This finding highlights the potential of exploring tuning-free approaches to achieve weak-to-strong generalization.\footnote{Code will be available at \url{https://github.com/THUNLP-MT/PANDA}}
\end{abstract}

\section{Introduction}
Large language models (LLMs) have shown exceptional performance across a broad spectrum of tasks via prompting~\citep{openai-gpt3,ouyang-etal-2021-dialogue,gpt4,touvron2023llama2,jiang2024mixtral}, suggesting they possess advanced general-purpose capabilities. Despite this, when tested on specific tasks, the effectiveness of these general-purpose LLMs often falls short of specialized state-of-the-art models~\citep{kocon2023chatgpt, zhong2023can, lin2023swiftsage}. Therefore, improving the domain-specific abilities of general-purpose LLMs remains a critical challenge.

\begin{figure}[t]
    \centering
    \includegraphics[width=\linewidth]{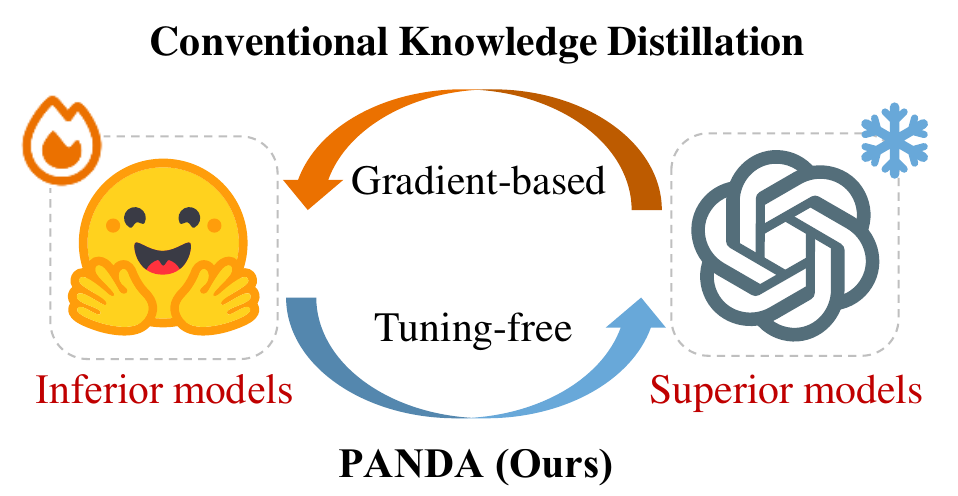}
    \caption{PANDA aims to enhance domain-specific capability of LLMs, which possess superior general capability, by learning from domain expert models that have inferior general capability with a tuning-free way. While conventional knowledge distillation usually leverage superior models to teach inferior models via gradient-based methods. The direction of the arrow represents the direction of knowledge transfer.}\label{fig:headPic}
\end{figure}

To address this challenge, one straightforward method is Knowledge Distillation~\citep[KD]{hinton2015distilling}, where the LLM acts as the student model and the domain expert model acts as the teacher model. However, it can be both resource-intensive and time-consuming. It might also inevitably reduce the original capabilities of the LLM, and in severe cases, could significantly impair them, as noted by \citet{qi2023fine}. Moreover, some of the most sophisticated LLMs~\citep{gpt4,geminiteam2023gemini} either do not offer a fine-tuning interface or provide only a limited set of fine-tuning options, rendering KD impractical. Therefore, KD is not an effective solution for this challenge.

Fortunately, tuning-free methods shed light on resolving the problem. Two notable classes of tuning-free methods are Retrieval-Augmentation-Generation (RAG) and self-reflection. RAG retrieves context information from a database during inference to enhance the prompt, thereby improving the performance of LLMs~\citep{gao2023retrieval}. However, previous RAG methods are not specifically designed to learn from expert models and do not fully utilize the capabilities of these models. On the other hand, self-reflection-based methods utilize LLMs to refine their outputs based on feedback from the environment~\citep{ shinn2023reflexion, wang2023learning}, ground-truth via self-reflection~\citep{yang2023large}, or feedback from domain expert models~\citep{ lin2023swiftsage, lu2023inference}. Despite their cost-efficiency due to being tuning-free, these methods often fall short of achieving the same level of performance as state-of-the-art (SOTA) models in specific domains~\citep{lin2023swiftsage, zhong2023can, lai2023chatgpt}. Additionally, the need to deploy an extra expert model during inference and the limited ability of LLMs to understand expert knowledge impose limitations on the potential performance improvement. Therefore, effectively leveraging expert models to fully equip LLMs with the capacity for specific tasks remains a critical issue to address.

To this end, we propose the \textbf{P}reference \textbf{A}daptation for E\textbf{n}hancing \textbf{D}omain-specific \textbf{A}bilities of LLMs (PANDA) method, a tuning-free approach for enhancing the domain-specific capabilities of LLMs. To achieve this, we initially task a domain expert model to infer on the training data and extract their output samples, which serve as expert preferences. Subsequently, we prompt the LLM to generate explanations (referred to as ``insights'') for the preferences of the expert model and gather these queries and insights into an ``insight pool''. This approach allows the LLM to develop a deeper understanding of the implicit knowledge  provided by the expert model  beyond mere surface behavior. During inference, PANDA retrieves the most relevant insights from the insight pool to assist in completing the current query, followed by standard inference using the retrieved insights. Compared to RAG, our work focuses on constructing a comprehensive knowledge base to better support LLMs in completing specific tasks. In contrast to self-reflection, our work emphasizes leveraging expert knowledge more effectively, rather than directly relying on the expert model, with the goal of achieving a higher ceiling performance. 

We conduct comprehensive experiments on interactive decision making and text classification tasks to evaluate the effectiveness of PANDA. Our experimental results demonstrate that PANDA enhances the domain-specific capability of LLMs across 16 tasks in ScienceWorld~\citep{wang2022scienceworld} and 7 tasks in TweetEval~\citep{barbieri2020tweeteval}. In summary, our contributions are as follows:

\begin{itemize} 
\item To the best of our knowledge, we are the first  to explore how to assist LLMs in effectively learning from expert models.

 \item We propose PANDA as a tuning-free approach that enhances the domain-specific capabilities of LLMs by adapting their preferences to align with the expert.

 \item Extensive experimental results demonstrate that PANDA significantly improves the domain-specific capabilities of LLMs across 16 tasks in ScienceWorld and 7 tasks in TweetEval. Particularly, LLMs with PANDA even outperforms the expert model that being learned on 4 tasks in ScienceWorld, revealing the potential of tuning-free approaches to achieve weak-to-strong generalization~\citep{burns2023weak}. 
\end{itemize}

\begin{figure*}[h]
    \centering
    \includegraphics[width=\linewidth]{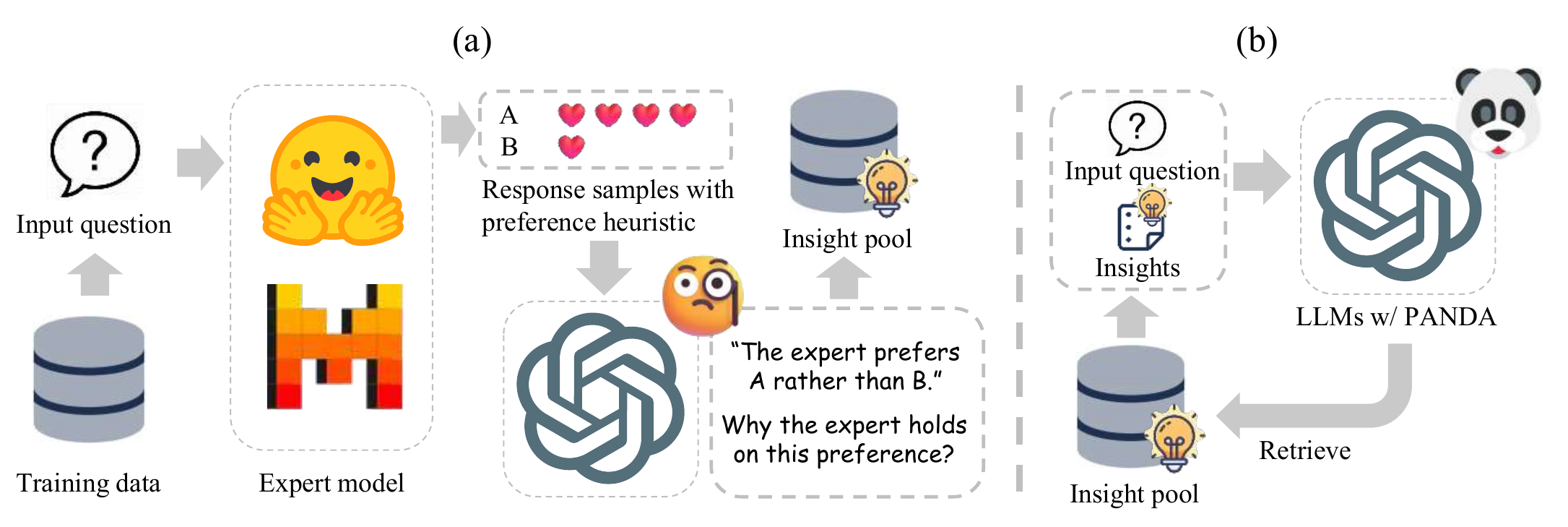}
    \caption{PANDA consists of two main stages: (a) the learning stage acquires insights from expert preferences and forms an insight pool; (b) the inference stage retrieves relevant insights from the insight pool and perform preference adaptation via in-context learning.}\label{fig:overview}
\end{figure*}

\section{Preference Adaptation for Enhancing Domain-Specific Abilities of LLMs}
\subsection{Overview}
As shown in Fig.~\ref{fig:overview}, PANDA consists of two main stages: the learning stage and the inference stage. The learning stage aims to capture the expert knowledge comprehensively. During this stage, the expert model is utilized to generate samples on the training dataset. PANDA then constructs preference pairs for each data instance based on specific preference heuristics. The preference heuristics may vary depending on the type of expert model being used. For a classification model, the preference rank can be determined based on the output logits distribution of each class, where higher logits may indicate that the expert model favors considering the specific query as belonging to that class..

Next, the LLM is prompted with the preference information to generate explanations. These explanations, derived from the training dataset, form the insight pool. During inference, the insight pool is utilized to retrieve relevant insights that can guide the LLM in adapting its preferences to align with those of the expert model. The goal is to enhance the domain-specific capability of the LLM by aligning its preferences with those of the expert model.

Meanwhile, it is worth noting that although PANDA and conventional KD differ significantly in terms of implementation techniques, they share a similar essence in each stage. We show discussion details in Section~\ref{sec:panda-vs-kd}.
\subsection{Learning from Expert Preferences}
When it comes to how to let LLMs learn from the domain expert models, a naive approach involves merely memorizing the behavior demonstrated by the expert models. A better approach involves not only memorizing the desired behavior but also considering the undesirable behaviors that the expert avoids in order to avoid making mistakes. Morever, the optimal approach goes beyond mere memorization of behavior; it also seeks to understand the underlying rationale behind actions of the expert. Drawing inspiration from this optimal learning process, we propose to learn from expert to boost the domain-specific performance of LLMs.

Inspired by the success of RLHF~\citep{ziegler2019fine}, which firstly trains a reward model using human preference data and then leverage it to perform reinforcement learning for LLMs to reach good alignment with humans. As Fig.~\ref{fig:overview}-(a) shows, we propose to learning from expert preferences as well since preferences information represents more comprehensive expert knowledge compared to soly behavior information. 

Leveraging the powerful language understanding and generation ability, we prompt LLMs to generate the explanation of the preferences of the expert, which is a process of learning expert knowledge. The simplified prompt template is as follows:
\begin{tcolorbox}[colback=gray!5!white,colframe=gray!75!black,title=Prompt for PANDA-Learning]
    \texttt{\{Query\}}
    
    The expert prefer \texttt{\{A\}} rather than \texttt{\{B\}}. Explain the reason why the expert holds on this preferences.
\end{tcolorbox}
Note that \texttt{A} and \texttt{B} here represent a pair of responses from the expert and its confidence in \texttt{A} is greater than it in \texttt{B}, which indicates the expert has a stronger preference for responding with \texttt{A} rather than \texttt{B} when presented with the \texttt{Query}. We refer to the response of LLMs to the prompt for learning as \texttt{Expert Insight}, which is then utilized to adapt the preferences of the LLMs towards better performance in completing the specific task. 

\subsection{Preference Adaptation during Inference}
Motivated by the intuition that strategies employed for similar problems often result in mutual benefits, we consider the problems with similar context as similar problems and insights from similar problems are considered relevant. As illustrated in Fig.~\ref{fig:overview}-(b), when encountering a new query, PANDA firstly retrieves related insights from the insight pool using the current query context as the retrieval key. In contrast to self-reflection-based methods, LLMs equipped with PANDA can achieve improved performance without the need for trial-and-error, thereby reducing the potential interaction cost.

Benefiting from the in-context learning ability of LLMs, we then incorporate the retrieved insights as part of the prompt, as described below:
\begin{tcolorbox}[colback=gray!5!white,colframe=gray!75!black,title=Prompt for PANDA-Inference]
    \texttt{\{prompt Context\}}
    
    These are some insights that may be helpful for you to improve success rate:
    
    \texttt{\{retrieved insights\}}
    
    \texttt{\{prompt Context\}}
\end{tcolorbox}
Compared to vanilla RAG-based methods that retrieve in-context examples, PANDA leverages insights from the expert preferences generated by LLMs, resulting in improved generalization.

\begin{table*}[h]
    \centering\small
    \begin{tabular}{l|c|cc|cc|cc}
    \toprule
    \textbf{Task}     & \textbf{Expert} & \textbf{ReAct}         & \textbf{ReAct$_{p}$}  & \textbf{Reflexion} & \textbf{Reflexion$_p$} & \textbf{SayCan} & \textbf{SayCan$_p$}  \\ \midrule 
    Task1-1  & 44.8   & \hphantom{0}0.0           & \hphantom{0}0.2   & \hphantom{0}0.0    &  \hphantom{0}0.1            &  \hphantom{0}0.7      &   \hphantom{0}1.0             \\ 
    Task1-4  & 30.7   & \hphantom{0}0.1 & \hphantom{0}0.0            & \hphantom{0}0.0    &  \hphantom{0}0.0            &  \hphantom{0}2.0      &   \hphantom{0}1.8             \\ 
    Task2-1  & \hphantom{0}8.7    & 11.3 & \hphantom{0}9.6            & 22.2   &  \hphantom{0}9.2            &  27.9     &   27.2             \\ 
    Task2-3  & \hphantom{0}5.8    & 33.3          & \underline{37.7}  & 73.8   &  \underline{77.6}           &  29.5     &   26.3            \\ 
    Task3-1  & 73.8   & 15.4          & \underline{22.8}           & 27.4   &  19.6           &  32.0     &   25.0            \\ 
    Task3-4  & 72.0   & 64.8          &  \underline{\textbf{76.3}} & 85.1   &  75.5           &  40.7     &   \underline{46.7}            \\ 
    Task4-1  & 100.0~~  & 15.1          & \underline{26.6}           & 16.4   &  \underline{28.4}           &  10.7     &   \underline{14.3}            \\ 
    Task4-2  & 96.7   & 64.1          & \underline{72.5}           & 81.4   &  \underline{85.0}           &  79.8     &   \underline{80.7}            \\ 
    Task5-1  & 28.5   & \hphantom{0}3.7           & \hphantom{0}\underline{5.5}            & \hphantom{0}6.0    &  \hphantom{0}6.3           &  \hphantom{0}8.3      &   \hphantom{0}8.4            \\ 
    Task5-2  & 17.0   & 36.2 & 23.7           & 52.5   &  49.4           &  42.8     &   \underline{50.1}             \\ 
    Task6-1  & 22.9   & 18.0          & \underline{\textbf{26.4}}  & 29.3   &  27.3           &  17.4     &   \underline{18.1}            \\ 
    Task6-3  & 13.7   & \hphantom{0}7.2           & \underline{10.4}  & 10.4   &  \underline{11.5}           &  \hphantom{0}7.8      &   \hphantom{0}\underline{8.6}            \\ 
    Task7-1  & 85.0   & 50.0          & \underline{\textbf{95.5}}  & 82.5   &  72.0           &  62.0     &   54.0             \\ 
    Task7-3  & 69.9   & 46.4          & \underline{\textbf{81.0}}  & 61.4   &  \underline{\textbf{75.2}}           &  50.6     &   48.8            \\ 
    Task8-1  & \hphantom{0}8.0    & \hphantom{0}4.0           & \hphantom{0}\underline{7.0}   & \hphantom{0}6.2    &  \hphantom{0}6.4            &  12.6     &   12.8            \\ 
    Task8-2  & 36.6   & \hphantom{0}0.0           & \hphantom{0}\underline{2.3}   & \hphantom{0}2.4    &  \hphantom{0}\underline{6.2}            &  \hphantom{0}9.1      &   \hphantom{0}9.2            \\ 
    Task9-2  & 41.5   & 16.5          & \underline{21.7}           & 31.7   &  \underline{38.5}           &  21.6     &   \underline{24.1}            \\ 
    Task9-3  & 66.5   & 10.0          & \underline{12.8}  & 14.4   &  \underline{18.9}           &  13.8     &   \underline{18.0}            \\ 
    Task10-1 & 16.9   & 21.7          & \underline{38.9}           & 44.6   &  39.4           &  29.9     &   \underline{33.5}           \\ 
    Task10-2 & 17.0   & \hphantom{0}0.2           & \hphantom{0}\underline{2.1}            & \hphantom{0}4.3    &  \underline{10.2}           &  25.3     &   \underline{32.4}            \\ \midrule
    \#n improvement & / & / & 16 &/ &9 & /& 10 \\
    \bottomrule
\end{tabular}

    \caption{Results on ScienceWorld. The expert model is a fine-tuned \texttt{flan-t5-large}. ReAct$_p$, Reflexion$_p$ and SayCan$_p$ represent the methods employed with PANDA. The term ``\#n improvement'' refers to the count of tasks achieving improved result when implementing PANDA. PANDA boosts ReAct to achieve ``weak-to-strong generalization'' on 4 tasks (7-3, 7-1, 6-1, 3-4). We mark scores that achieve ``weak-to-strong generalization'' in \textbf{bold}, and scores that show improvement with PANDA in \underline{underline}.}
    \label{tab:overallResSW}
    \end{table*}

\begin{table*}[]
\centering\small
\begin{tabular}{l|ccccccc}
\toprule
          & \textbf{Emoji} & \textbf{Emotion}       & \textbf{Offensive}     & \textbf{Abortion} & \textbf{Atheism} & \textbf{Climate} & \textbf{Sentiment}     \\ \midrule
Expert    & 29.7  & 80.7         & 81.6          & 52.2   & 72.0             & 54.3           & 71.6          \\ \midrule
Zero-Shot & \hphantom{0}8.4   & 65.7          & 69.6 & 55.6            & 35.4           & 63.6           & 63.0          \\
w/ PANDA  & \underline{10.6}  & 61.1          & 68.9          & \underline{60.1}            & \underline{48.3}           & \underline{65.2}           & \underline{65.8}          \\ \midrule
Few-Shot  & \hphantom{0}8.8   & 74.1          & 74.2          & 58.4            & 25.6           & 50.3           & 65.9          \\
w/ PANDA  & \underline{10.3}  & \underline{77.9} & 72.9          & \underline{61.1}            & \underline{55.4}  & \underline{69.1}  & \underline{67.1} \\ \midrule
ZS-CoT    & 15.1  & 53.4          & 44.9          & 39.9            & 23.7           & 48.9           & 48.6          \\
w/ PANDA  & \underline{20.1}  & \underline{61.1}          & \underline{47.1}          & \underline{47.7}            & \underline{42.5}           & 47.5           & \underline{50.1}          \\ \midrule
FS-CoT    & 18.3  & 52.6          & 48.1          & 61.7            & 22.4           & 58.3           & 48.4          \\
w/ PANDA  & \underline{19.0}    & \underline{57.5}          & \underline{48.8}          & \underline{66.3}   & \underline{28.2}           & \underline{63.2}           & 48.5          \\ \bottomrule
\end{tabular}
\caption{Main results on 7 tasks from TweetEval. The expert model is fine-tuned \texttt{RoBERTa-base} models. The introduction of PANDA significantly enhances the performance of GPT-3.5-turbo across almost all tasks in four different settings: zero-shot, few-shot, zero-shot Chain-of-Thought (ZS-CoT), and few-shot Chain-of-Thought (FS-CoT). We mark scores that show improvement with PANDA in \underline{underline}.}
    \label{tab:mainResTweet}
\end{table*}

\section{Experiments}
To validate effectiveness of PANDA, We evaluate it on interactive decision making and text classification tasks. In all experiments, we utilize SentenceBERT~\citep{reimers2019sentence} to calculate sentence embeddings and retrieve relevant insights base on cosine similarity. We set the number of retrieved insights to 1 in ScienceWorld and 6 in TweetEval. All of the experiments take \texttt{gpt-3.5-turbo-1106} as the LLM in PANDA.
\subsection{Interactive Decision Making}
\paragraph{Benchmark.} In order to evaluate the effectiveness of PANDA in interactive decision-making tasks, we selected ScienceWorld~\citep{wang2022scienceworld} as the benchmark, which is an 
interactive text simulation of a laboratory environment, consisting of 10 sub-rooms such as Green House, Foundry, and Workshop. It provides a rich set of actions, with approximately 200k possible actions per step, including actions like moving to the greenhouse or picking up a jug, thereby posing challenges for LLMs in terms of their capabilities in reasoning, planning and gounding to the embodied environmet. 
The ScienceWorld benchmark encompasses 30 tasks across 10 distinct classes. Each task exhibits significant variations, including differences in environment configurations (e.g., objects present in different rooms), specific task goals, and scientific domains (e.g., growing a plant, measuring the boiling point of substances). The number of steps required to complete each task varies, ranging from under 10 steps to over 100 steps.
To ensure fairness and manage cost constraints, we randomly sampled 2 tasks from each task class, resulting in a total of 20 tasks for our evaluation.

\paragraph{Baselines.}  We compare PANDA with three baselines: ReAct~\citep{yao2022react}, Reflexion~\citep{shinn2023reflexion} and SayCan~\citep{ahn2022can}, which have been proposed for grounding LLMs in agent tasks and enhancing their performance in complex interactive tasks. 
ReAct combines reasoning and acting in LLMs to enhance their ability to solve complex interactive tasks.
Reflexion prompts LLMs to generate reflections on their failure trajectory, providing verbal reinforcement feedback to aid in their self-improvement.
SayCan leverages a value function that implicitly serves as a policy, utilizing grounding information about the environment to guide LLMs in task completion.
We adopted the same implementation details as~\citep{lin2023swiftsage}, with the exception that all our experiments utilize the \texttt{gpt-3.5-turbo-1106} model. Furthermore, while maintaining rationality, we extended the original observation from the environment by concatenating it with information about the current rooms and inventory.

As the expert model, we utilize a fine-tuned \texttt{flan-t5-}\texttt{large}~\citep{lin2023swiftsage}, which is the state-of-the-art single-model on ScienceWorld. It was fine-tuned using a multi-hop behavior cloning strategy, and the original dataset was downsampled to achieve good multi-task performance.
\begin{tcolorbox}[colback=gray!5!white,colframe=gray!75!black,title=\small{Prompt of PANDA-Inference for ScienceWorld}]
    \texttt{\{Init Prompt\}}
    
    These are some insights that may be helpful for you to improve the success rate:
    
    \texttt{\{Retrieved Insights\}}

    \texttt{\{Current Trajectory\}}
\end{tcolorbox}
For PANDA, we performed beam search and selected the top-2 responses from the expert model to construct the preference pair and learning prompt. The observation of the environment served as the retrieval key for the insight pool, inspired by the intuition that strategies employed for similar problems often yield mutual benefits.

\paragraph{Results.} Since PANDA is agnostic to the underlying agent, we evaluate PANDA in comparison to each baseline. As shown in Table~\ref{tab:overallResSW}, even powerful agents like ReAct and SayCan fail to surpass the expert model. PANDA improves upon the ReAct baseline in 16 out of 20 tasks, the SayCan baseline in 10 tasks and even improves Reflexion in 9 tasks out of 20 tasks, which is an extremely strong baseline agent. Notably, ReAct with PANDA achieves a performance that surpasses the expert model in 4 tasks, demonstrating a similar phenomenon of \textit{weak-to-strong generalization} observed in~\citep{burns2023weak}. Moreover, PANDA achieves these results without requiring any tuning, making it a promising approach towards achieving superalignment with LLMs. In Task 5-2 and Task 2-1, where ReAct outperforms the expert model, PANDA, in its effort to align the preferences of ReAct with the expert model, leads to a degradation in the performance of ReAct. It suggests that PANDA adjusts the preferences of ReAct to better conform to the preferences of the expert model, even in cases where the expert model may not perform well.

\subsection{Text Classification}
\paragraph{Benchmark.} TweetEval~\citep{barbieri2020tweeteval} is a benchmark that consists of 11 heterogeneous Twitter-specific classification tasks, including Sentiment, Emoji, Emotion, Hate, Irony, Offensive and Stance. These tasks encompass both pragmatic and semantic aspects of tweet classification. It is worth noting that the distribution of samples across different classes in TweetEval tasks can be imbalanced. In our evaluation, we utilized the macro-averaged F1 score as the performance metric. Since PANDA aims to enhance the domain-specific capabilities of LLMs through learning from domain experts, in this section we primarily focus on presenting the results of tasks where \emph{the expert model consistently outperforms LLMs}. This allows us to showcase the potential of PANDA in addressing the performance gap between LLMs and domain-specific fine-tuned models in these challenging tasks.

\paragraph{Baselines.} As LLMs have strong instruction-following ability, we take zero-shot as a baseline and few-shot with 3 exemplars serving as a stronger baseline. We additionally test PANDA upon Chain-of-Thought~\citep{wei2022chain} within zero-shot and few-shot settings, which output intermediate rationales to boost reasonning capability of LLMs.
\begin{tcolorbox}[colback=gray!5!white,colframe=gray!75!black,title=\small{Prompt of PANDA-Inference for TweetEval}]
    These are some insights that may be helpful for you to improve the success rate:
    
    \texttt{\{Retrieved Insights\}}

    \texttt{\{Zero-shot/ Few-shot/ CoT Prompt\}}
\end{tcolorbox}
We utilize the fine-tuned \texttt{RoBERTa-base} models~\footnote{https://huggingface.co/cardiffnlp/roberta-base-emotion}, which has been tuned on the training data for each task, as our domain expert model for GPT-3.5-turbo to learn from. We take the top-2 classes as the preference pairs data to learn from. To facilitate this learning process, we extract preference pairs data by sampling the top two classes for each task.

\paragraph{Results.}\label{sec:tweetEvalRes} As indicated in Table~\ref{tab:mainResTweet}, GPT-3.5-turbo falls short of achieving the same level of performance as the fine-tuned \texttt{roberta-base} models on 7 tasks within the TweetEval dataset. Even when employing powerful prompt techniques like few-shot or Chain-of-Thought, the performance of the fine-tuned \texttt{RoBERTa-base} models still surpasses that of GPT-3.5-turbo on 5 tasks.
However, the introduction of PANDA significantly enhances the performance of GPT-3.5-turbo across almost all tasks in four different settings: zero-shot, few-shot, zero-shot Chain-of-Thought (ZS-CoT), and few-shot Chain-of-Thought (FS-CoT). It is noteworthy that PANDA improves the performance of GPT-3.5-turbo on the climate and abortion stance classification tasks, where the expert model initially performed worse than GPT-3.5-turbo. This implies that PANDA not only incorporates the strengths of other models but also extracts valuable knowledge from their weaknesses, which can aid in self-improvement.

\begin{table*}[h]
    \centering\small
\begin{tabular}{l|c|c|ccc|cc}
\toprule
\textbf{Task}     & \textbf{Expert} & \textbf{ReAct}         & \textbf{PANDA 3}       & \textbf{PANDA 2}        & \textbf{PANDA 1}        & \textbf{Raw 1} & \textbf{Raw 2}          \\ \midrule
Task1-4  & 30.7   & \hphantom{0}0.1 & \hphantom{0}0.0          & \hphantom{0}0.0           & \hphantom{0}0.0           & \hphantom{0}0.0   & \hphantom{0}0.0           \\ 
Task3-4  & 72.0   & 54.4          & \textbf{79.9}         & 76.3          & 73.9 & 67.6  & 61.4          \\ 
Task4-1  & 100.0~~  & 15.1          & 28.5         & 26.6          & \textbf{36.2} & 22.8  & 33.4          \\ 
Task4-2  & 96.7   & 64.1          & 73.2         & 72.5          & 74.7          & 75.2  & \textbf{76.7} \\ 
Task5-1  & 28.5   & \hphantom{0}3.7           & \hphantom{0}\textbf{6.0} & \hphantom{0}5.5           & \hphantom{0}5.9           & \hphantom{0}5.3   & \hphantom{0}5.9           \\ 
Task6-3  & 13.7   & \hphantom{0}7.2           & \hphantom{0}9.7          & \textbf{10.4} & \hphantom{0}9.2           & \hphantom{0}9.4   & \hphantom{0}9.4           \\ 
Task7-3  & 69.9   & 46.4          & 76.7         & \textbf{81.0} & 77.3          & 68.1  & 59.8          \\ 
Task8-2  & 36.6   & \hphantom{0}0.0           & \hphantom{0}1.2          & \hphantom{0}\textbf{2.3}  & \hphantom{0}1.1           & \hphantom{0}1.5  & \hphantom{0}0.6          \\ 
Task9-3  & 66.5   & 10.0          & \textbf{29.4}         & 12.8 & 12.1          & 10.2  & 10.0          \\ 
Task10-2 & 17.0   & \hphantom{0}0.2           & \hphantom{0}4.1          & \hphantom{0}2.1           & \hphantom{0}\textbf{6.4}  & \hphantom{0}0.1   & \hphantom{0}2.2          \\ 
\bottomrule
\end{tabular}

 \caption{Ablation results on ScienceWorld. ``PANDA $n$'' means learning from top-$n$ responses generated from the expert model. ``Raw 1'' ablates PANDA insights to raw behavior of expert model (e.g., ``the expert prefers A''). ``Raw 2'' ablates PANDA insights to raw preferences of expert model (e.g., ``the expert prefers A rather than B''). We mark the best score except for the expert model in \textbf{bold} for each task.}
    \label{tab:ablationSWRes}
\end{table*}

\section{Ablation and Analysis}
\subsection{Ablation Study}
\paragraph{PANDA elicits generalizable knowledge from preferences data.} We conducted an ablation study on 10 tasks of ScienceWorld benchmark  to examine the effectiveness of PANDA. As shown in Table~\ref{tab:ablationSWRes}, PANDA consistently outperforms the ablation cases of ``Raw 1'' or ``Raw 2'', which consider raw preferences (e.g., ``the expert prefers A rather than B'') without additional reasoning. This demonstrates that the learning process of PANDA, which involves gathering insights by reasoning about the preferences of the expert in specific contexts, leads to improved performance, except for only one exceptional case.
We further explore the impact of incorporating preference data from the expert model by conducting ablation experiments on ``PANDA n'', where n represents the learning process utilizing preference data of the top-$n$ responses. We observed that ``PANDA 2'' achieves better performance than ``PANDA 1''. Although the performance improvement of ``PANDA 2'' may appear to be marginal, our additional results on 20 tasks in Appendix~\ref{res:addSW} further validate this conclusion. This indicates that learning from data that contains more comprehensive information about the preferences of the expert elicits more useful insights, which can effectively enhance the performance of language models.
Ideally, having more preferences information from the expert would elicit more expert knowledge. However, our experimental results show no significant difference between ``PANDA 3'' and ``PANDA 2''. This could be attributed to the difficulty for language models to reason and extract informative insights from multi-hop questions. Therefore, we decided to leverage ``PANDA 2'' for all our main experiments, as it demonstrated strong performance and captured valuable insights from the preferences data.
\begin{figure}[t]
    \centering\scalebox{1.0}{
    \includegraphics[width=\linewidth]{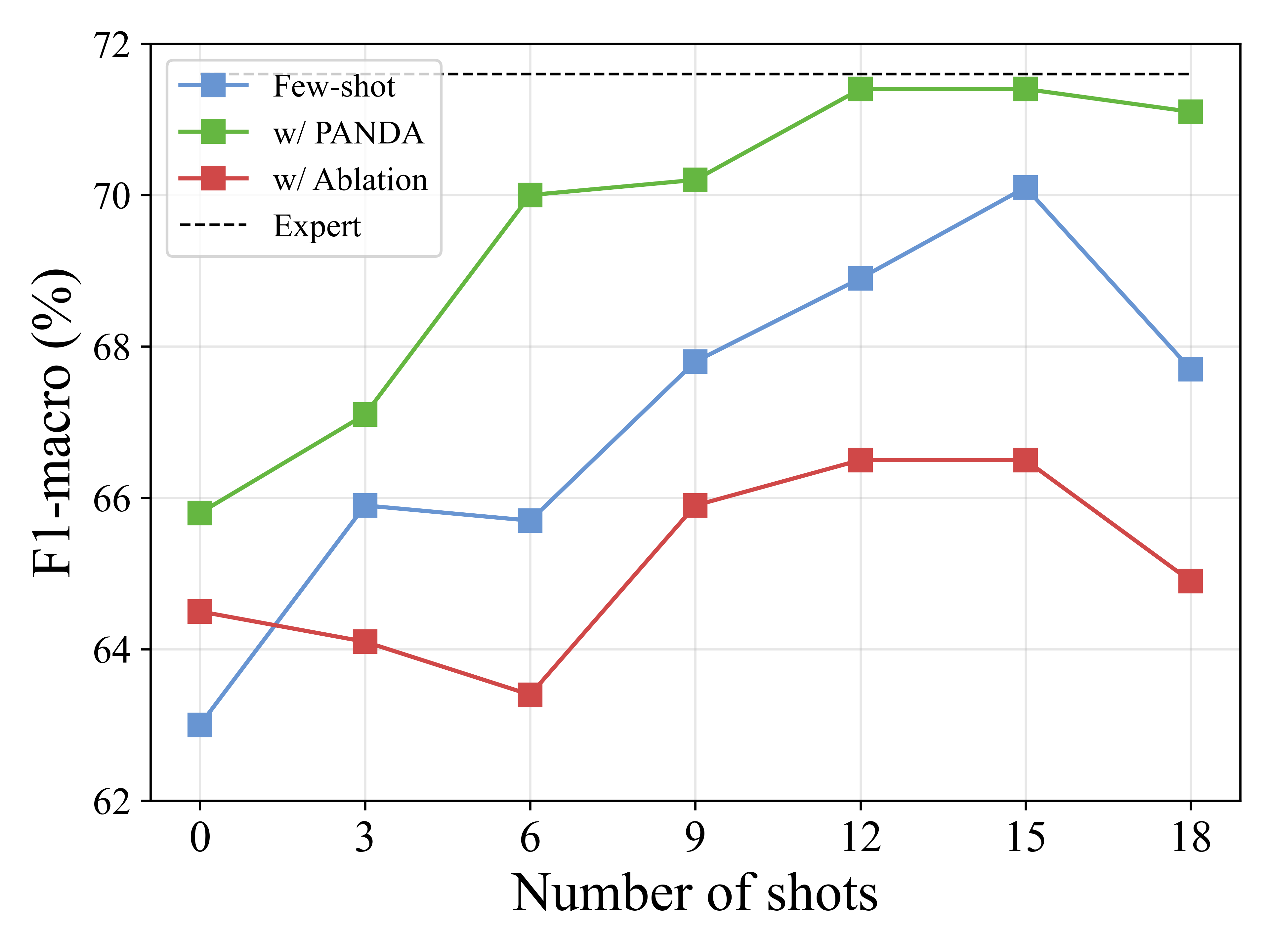}}
    \caption{Ablation results on sentiment classification. For ablation study, we replace the retrieved insights with the corresponding few-shot examples (``w/ Ablation''), whose label is provided by the expert model.}\label{fig:ablaSenti}
\end{figure}

\paragraph{Improvement derived from PANDA is not the substitution of more shots in the prompt.} We conducted an additional experiment on the sentiment classification task to demonstrate that PANDA can surpass the performance ceiling of few-shot learning. In Fig.~\ref{fig:ablaSenti}, we observe that the few-shot baseline achieves its highest performance when the number of shots is set to 15. Interestingly, PANDA consistently outperforms the few-shot baseline across all settings of shot numbers, indicating that improvement of PANDA cannot be substituted by adding more few-shot examples. To further investigate this, we performed an ablation where we replaced the retrieved insights with the corresponding few-shot examples (``w/ Ablation''), whose label is provided by the expert model. The results show that the ablation consistently performs worse than the baseline when the number of shots is greater than zero. This degradation in performance may be due to the pseudo labels generated by the expert model, as the expert model is not perfect and its predictions may not always be accurate. In contrast, PANDA achieves consistent improvement by leveraging summary insights from the preferences of the expert. These insights better represent the understanding of the expert for the specific task compared to raw prediction results, leading to better generalization and improved performance.

\begin{table}[]
\centering\small
\begin{tabular}{l|l|ll}
\toprule
\multicolumn{1}{c|}{\textbf{Task-setting}} & \multicolumn{1}{c|}{\textbf{ReAct}} & \multicolumn{1}{c}{\textbf{ReAct$_p$}} & \multicolumn{1}{c}{\textbf{ReAct$_a$}} \\ \midrule
Task1-1→1-4                                & \hphantom{0}0.1                                 & \hphantom{0}0.0                                                 & \hphantom{0}0.0                                                           \\
Task2-1→2-3                                & 33.3                                & 32.3                                              & 11.1                                                        \\
Task3-1→3-4                                & 64.8                                & \textbf{76.3}                                              & 52.0                                                          \\
Task4-1→4-2                                & 64.1                                & \textbf{75.4}                                              & \textbf{76.8}                                                        \\
Task5-1→5-2                                & 36.2                                & 20.2                                              & 22.3                                                        \\
Task6-1→6-3                                & \hphantom{0}7.2                                 & \hphantom{0}\textbf{8.8}                                               & \hphantom{0}7.5                                                         \\
Task7-1→7-3                                & 46.4                                & \textbf{71.5}                                              & \textbf{58.7}                                                        \\
Task8-1→8-2                                & \hphantom{0}0.0                                   & \hphantom{0}\textbf{2.4}                                               & \hphantom{0}\textbf{1.5}                                                         \\
Task9-2→9-3                                & 10.0                                  & 10.3                                              & \hphantom{0}1.1                                                         \\
Task10-1→10-2                              & \hphantom{0}0.2                                 & \hphantom{0}\textbf{2.5}                                               & \hphantom{0}0.3                                                         \\ \bottomrule
\end{tabular}
\caption{ Cross task transference results on ScienceWorld. ``Task1-1→1-4'' indicates that we utilize PANDA to acquire expert knowledge in Task 1-1 and subsequently evaluate it in Task 1-4. This pattern is consistent across other tasks. We mark the positive transference in bold. Positive transference means ReAct$_{p}$ (learning w/ PANDA) or ReAct$_a$ (learning w/ ablated PANDA) outperforms ``ReAct'' by at least 1.0.}
\label{tab:TransResSW}
\end{table}
\begin{table*}[]
\small
\begin{floatrow}

\centering
\capbtabbox{
\begin{tabular}{l|c}
\toprule 
\multicolumn{1}{c|}{\textbf{Method}} & \multicolumn{1}{c}{\textbf{Emotion→Sentiment}} \\ \midrule
Zero-Shot                            & 63.0                                           \\
w/ PANDA                             & \textbf{64.6}                                           \\
w/ Ablation                          & 57.4                                           \\ \midrule
Few-Shot                             & 65.9                                           \\
w/ PANDA                             & \textbf{67.7}                                           \\
w/ Ablation                          & 65.4                                           \\ \bottomrule
\end{tabular}}{
\caption{Cross task transference results on TweetEval, with emotion classification serving as the source task and sentiment classification as the target task.}
\label{tab:TransResTweet}}

\capbtabbox{
\begin{tabular}{l|c|cc|cc}
\toprule
\textbf{\#N shot} & \textbf{Baseline} & \textbf{PANDA$_p$} & \textbf{Ablation$_p$} & \textbf{PANDA$_{gt}$} & \textbf{Ablation$_{gt}$} \\ \midrule
0                 & 63.0              & \textbf{65.8}        & 64.5                    & \textbf{64.4}         & 62.5                     \\
3                 & 65.9              & \textbf{67.1}        & 64.1                    & \textbf{67.8}         & 64.0                     \\
6                 & 65.7              & \textbf{70.0}        & 63.4                    & \textbf{69.6}         & 64.3                     \\
9                 & 67.8              & \textbf{70.2}        & 65.9                    & \textbf{70.6}         & 67.0                     \\
12                & 68.9              & \textbf{71.4}        & 66.5                    & \textbf{71.2}         & 67.2                     \\
15                & 70.1              & \textbf{71.4}        & 66.5                    & \textbf{70.8}         & 66.5                     \\
18                & 67.7              & \textbf{71.1}        & 64.9                    & \textbf{70.1}         & 65.5                     \\ \bottomrule
\end{tabular}}{
\caption{Ablation results on sentiment classification. For ablation study, we replace the retrieved insights with the corresponding few-shot examples (Ablation$_p$ and Ablation$_{gt}$). We mark the improved score when implementing PANDA in bold.}
\label{tab:ResTweetGT}
}
\end{floatrow}
\end{table*}
\subsection{Generalization in Out-of-Domain Settings}
To assess the generalizability of PANDA in out-of-domain settings, we conduct experiments on ScienceWorld and TweetEval. Specifically, we evaluate the transference performance of PANDA in ScienceWorld by learning on one task and testing it on another task within the same topic (Table \ref{tab:TransResSW}). Tasks within the same topic are related but vary a lot. For instance, in task 5, we let PANDA learn on Task 5-1 (grow-plant) and test it on Task 5-2 (grow-fruit). As our ablation analysis (ReAct$_a$), we remove the insight pool from PANDA and use raw preference data instead. Our experimental results on ScienceWorld demonstrate that PANDA achieved positive transfer in 7 out of 10 tasks. This highlights the superior generalization capability of PANDA in out-of-domain scenarios. For comparison, the ablated setting only showed positive transfer in 3 out of 10 tasks, underscoring the efficacy of learning from expert preferences proposed in PANDA.

Furthermore, we conducted cross-task transference experiments on TweetEval (Table \ref{tab:TransResTweet}). To be specific, our experiments involved transferring from task-emotion to task-sentiment. Our experimental results show that in the ``emotion→sentiment'' setting, PANDA exhibits notable positive transference in both zero-shot and few-shot scenarios, while the ablations (``w/ Ablation'') show significant negative transference, consistent with the findings in ScienceWorld. These results collectively underscore the cross-task transference efficacy of PANDA.

\subsection{Performance in Presence of Ground-Truth}
We firstly conduct supplementary experiments to assess the performance of PANDA when utilizing ground truth labels instead of expert predictions. As depicted in Table \ref{tab:ResTweetGT}, the results indicate an enhancement in ablation performance with the utilization of ground truth data (``Ablation$_{gt}$'' vs. ``Ablation$_p$''). Despite this improvement, PANDA maintains its superiority over the ablation method, thereby affirming the effectiveness of PANDA in distilling generalizable knowledge.

\begin{table*}[]\centering\small
\begin{tabular}{l|cccccc}

\toprule
\textbf{\#N shot} & \textbf{Baseline} & \textbf{TA=25\%} & \textbf{TA=50\%} & \textbf{TA=70\%} & \textbf{TA=82.4\%} & \textbf{TA=100\%} \\ \midrule
0                 & 63.0              & 62.7             & 64.6             & 64.0             & 65.8               & 64.4              \\
3                 & 65.9              & 65.6             & 68.5             & 67.1             & 67.1               & 67.8              \\
6                 & 65.7              & 68.5             & 68.5             & 68.8             & 70.0               & 69.6              \\
9                 & 67.8              & 69.4             & 69.3             & 69.1             & 70.2               & 70.6              \\
12                & 68.9              & 68.2             & 69.2             & 69.1             & 71.4               & 71.2              \\
15                & 70.1              & 68.7             & 69.8             & 69.4             & 71.4               & 70.8              \\
18                & 67.7              & 69.7             & 70.7             & 69.3             & 71.1               & 70.1              \\ \bottomrule
\end{tabular}
\caption{Results for PANDA with varied training data quality. The baseline denotes the vanilla N-shot. TA denotes the accuracy of the training data (reflecting its quality). TA=82.4\% corresponds to labels predicted by the expert model, while TA=100\% represents ground truth labels.}
\label{tab:TweetGTwTA}
\end{table*}
We also note that implementing PANDA using ground truth labels does not exhibit significant improvement compared to using the predictions of an expert model. To investigate this further, we conduct additional experiments by manually synthesizing training data with different labels quality, through flipping some of the ground truth labels. The experimental results are presented in Table \ref{tab:TweetGTwTA}. It is evident that irrespective of the baseline N-shot setting, the efficacy of PANDA demonstrates an increasing trend with the improvement in the quality of the training data.

Additionally, it is noteworthy that while training data with higher-quality does indeed enhance the performance of PANDA, PANDA exhibits relative robustness to variations in training data quality. This is evident from the fact that PANDA still achieves improvements compared to the baseline even when the accuracy of the training data is as low as 25\%, which further demonstrate the superiority of PANDA for distilling generalizable knowledge.

\section{Related Work}
\paragraph{Knowledge Distillation.} Knowledge Distillation~\citep[KD]{hinton2015distilling} is a learning paradigm that involves training a inferior model to learn from a superior model in order to enhance the capabilities of the inferior model. Some work~\citep{gu2023knowledge} proposes reverse KL divergence for generative language models to alleviating the problem caused by the distribution gap between teacher and student model. There is also work~\citep{hsieh2023distilling} leveraging LLMs to generate intermediate rationales that facilitates small language model to better handling the knowledge from large teacher model. Morever,~\citep{huang2022context} implements Meta In-context Tuning and Multitask In-context Tuning to transfer few-shot learning ability of pretrained language models to small language models. However, all of these methods are not only gradient-based that cannot be used on close-source models but also computation-intensive. As a contrast, our proposed PANDA leveraging the strong language understanding ability of LLMs is tuning-free. Meanwhile, most of works in KD leverage inferior models to learn from superior models, our work explores a new paradigm that leverages LLMs to learn from domain-specific expert models that may be much smaller than LLMs.
\paragraph{Boosting LLMs with Small Language Models.} Lots of works leverage small language model to boost the performance of LLMs. Some methods focus on extending the task-specific capabilities of LLMs by utilizing small language models to enhance in-context examples~\citep{xu2023small} or by incorporating post-process modules~\citep{vernikos2023small}. Retroformer~\citep{yao2023retroformer} employs reinforcement learning to fine-tune a language model as a reflection-specific module, generating high-quality reflections in a loop that involves frozen LLMs. Another approach~\citep{lu2023inference} employs small language models as decoding-time tailors for LLMs, aiming to improve task-specific performance. Some work~\citep{liu2024tuning} leverage the output distribution gap between pre- and post-tuning of small language models as a heuristic to adjust the output distribution of LLMs, improving task-specific performance. Although PANDA also aims to leverage small models to enhance LLMs, there is a significant distinction that PANDA enables LLMs to learn from small models directly with a tuning-free manner, similar to the weak-to-strong paradigm proposed by OpenAI~\citep{burns2023weak}, thus being more flexible.

\paragraph{Self-improvement of LLMs.} Building upon the strong foundational capabilities of LLMs, several approaches~\citep{chen2024self, yuan2024self, qiao2024autoact, aksitov2023rest} have been proposed to enhance LLMs without the need for human supervision. To be specific, SPIN~\citep{chen2024self} treats LLMs as both opponents and main players to train LLMs using a bootstrapping way. Some work~\citep{yuan2024self} proposes to enable LLMs to generate both their training data and reward signals, allowing them to fine-tune themselves using the generated data. There are also works that tune language models with trajectory data generated by their own to boost themselves~\citep{qiao2024autoact, aksitov2023rest}. Morever, weak-to-strong generalization proposed by~\citep{burns2023weak} is another potential paradigm for boosting LLMs capabilities, which shares similarities with PANDA. A key difference is that PANDA leverages LLMs to learn from small domain-expert models in a purely tuning-free manner.

\section{Conclusion}
We propose PANDA, a tuning-free method that aims to improve the task-specific capability of LLMs. In PANDA, we leverage LLMs to learn from preferences of expert model and form an insight pool during learning stage. At inference time, PANDA firstly retrieved relevant insights to current query and boost the performance of LLMs by adapting its preference to align with the expert model with the guidance of relevant insights. Through extensive experiments on 20 tasks in ScienceWorld and 11 tasks in TweetEval, we verify that PANDA improve LLMs in most cases. Notably, PANDA achieves the weak-to-strong generalization on 4 tasks in ScienceWorld based on ReAct, which highlights the potential of exploring tuning-free approaches to achieve weak-to-strong generalization.
\section*{Limitations}
PANDA has two main limitations. Firstly, it relies on the retrieval process, where relevant insights need to be retrieved from the insight pool during inference. As the complexity of the task increases, retrieving the appropriate insights becomes more challenging, potentially causing the retrieval stage to become a performance bottleneck for PANDA.
Secondly, PANDA is dependent on the powerful instruction-following capability of LLMs during the learning stage. This requirement restricts the usage of PANDA to LLMs with robust language understanding and instruction-following capabilities, limiting its applicability to open-sourced LLMs.

\section*{Acknowledgement}
This work is supported by the National Natural Science Foundation of China (No. 62276152, 61925601) and Institute Guo Qiang at Tsinghua University.
\bibliography{anthology,custom}
\bibliographystyle{acl_natbib}

\newpage
\appendix

\section{Additional Experimental Results}
\subsection{ScienceWorld}
We conduct additional experiments to validate the effectiveness of learning from preferences compared to learning from individual behavior. As shown in Table~\ref{tab:SW-one-vs-two}, PANDA with learning from preference pair outperforms PANDA with learning from individual behavior.
\label{res:addSW}

\subsection{TweetEval}
A result confusing us is that the ablation w/ ground truth is worse than the n-shot baseline in Table \ref{tab:ResTweetGT}. To further demonstrate the effecicy of ICL examplars to the ablation. We conduct experiments of ablations under the same training data configurations as Table \ref{tab:TweetGTwTA}. Results are as Table \ref{tab:tweetTAwAblation} shows. We can observe that, as the quality of training data increases, the performance of the ablation of PANDA shows an increasing trend. However, even with the presence of the groundtruth label, the ablations still underperforms the baselines. We atrribute this to the lack of sophisticated ICL tricks such as tuning the ordering of the exemplars~\cite{liu2022makes}, selecting exemplars according to the intrinsic knowledge of LLMs~\cite{lee2023crafting}. In comparison, PANDA consistently achieves improvement under different settings using vanilla RAG, further highlighting its superiority.

\section{Experimental Setup}
In all experiments, we take \texttt{gpt-3.5-turbo-1106} as the LLM in PANDA. Due to the observed variability in results within the ScienceWorld benchmark, even when the \texttt{temperature} is fixed at $0.0$, we conducted each experiment for 5 rounds and recorded the average score as the final measurement.
\begin{table}[h]
    \centering\small
    \begin{tabular}{l|c|ccc}
    \toprule
    \textbf{Task}     & \textbf{Expert} & \textbf{ReAct}         & \textbf{ReAct$_{p2}$}  & \textbf{ReAct$_{p1}$}  \\ \midrule 
    Task1-1  & 44.8   & \hphantom{0}0.0           & \hphantom{0}\textbf{0.2}   & \hphantom{0}0.1\\
    Task1-4  & 30.7   & \hphantom{0}\textbf{0.1} & \hphantom{0}0.0            & \hphantom{0}0.0\\ 
    Task2-1  & \hphantom{0}8.7    & \textbf{11.3} & \hphantom{0}9.6            & \hphantom{0}7.0\\ 
    Task2-3  & \hphantom{0}5.8    & 33.3          & \textbf{37.7}  & 17.0\\ 
    Task3-1  & 73.8   & 15.4          & 22.8           & \textbf{32.9}\\ 
    Task3-4  & 72.0   & 64.8          &  \textbf{76.3} & 73.9\\ 
    Task4-1  & 100.0~~  & 15.1          & 26.6           & \textbf{36.2}\\ 
    Task4-2  & 96.7   & 64.1          & 72.5           & \textbf{74.7}            \\ 
    Task5-1  & 28.5   & \hphantom{0}3.7           & \hphantom{0}5.5            & \hphantom{0}\textbf{5.9}\\ 
    Task5-2  & 17.0   & \textbf{36.2} & 23.7           & 22.3\\ 
    Task6-1  & 22.9   & 18.0          & \textbf{26.4}  & 21.7\\ 
    Task6-3  & 13.7   & \hphantom{0}7.2           & \textbf{10.4}  & \hphantom{0}9.2            \\ 
    Task7-1  & 85.0   & 50.0          & \textbf{95.5}  & 91.5 \\ 
    Task7-3  & 69.9   & 46.4          & \textbf{81.0}  & 77.3\\ 
    Task8-1  & \hphantom{0}8.0    & \hphantom{0}4.0           & \hphantom{0}\textbf{7.0}   & \hphantom{0}6.0\\ 
    Task8-2  & 36.6   & \hphantom{0}0.0           & \hphantom{0}\textbf{2.3}   & \hphantom{0}1.1\\ 
    Task9-2  & 41.5   & 16.5          & 21.7           & \textbf{23.3}\\ 
    Task9-3  & 66.5   & 10.0          & \textbf{12.8}  & 12.1\\ 
    Task10-1 & 16.9   & 21.7          & 38.9           & \textbf{43.5}\\ 
    Task10-2 & 17.0   & \hphantom{0}0.2           & \hphantom{0}2.1            & \hphantom{0}\textbf{6.4} \\ 
    \bottomrule
\end{tabular}

    \caption{Ablation results on ScienceWorld. ReAct$_{p2}$ and ReAct$_{p1}$ represents PANDA learning from preference pair and individual behavior respectively.  We mark the best score except for the expert model in \textbf{bold} for each task.}
    \label{tab:SW-one-vs-two}
    \end{table}

\begin{table*}[]
\centering\small
\begin{tabular}{l|cccccc}
\toprule
\textbf{\#N shot} & \textbf{Baseline} & \textbf{TA=25\%} & \textbf{TA=50\%} & \textbf{TA=70\%} & \textbf{TA=82.4\%} & \textbf{TA=100\%} \\ \midrule
0                 & 63.0              & 62.0             & 62.8             & 62.6             & 64.5               & 62.5              \\
3                 & 65.9              & 62.3             & 63.4             & 62.6             & 64.1               & 64.0              \\
6                 & 65.7              & 62.2             & 63.7             & 62.9             & 63.4               & 64.3              \\
9                 & 67.8              & 66.3             & 66.9             & 66.1             & 65.9               & 67.0              \\
12                & 68.9              & 65.1             & 65.5             & 65.5             & 66.5               & 67.2              \\
15                & 70.1              & 65.9             & 65.4             & 66.3             & 66.5               & 66.5              \\
18                & 67.7              & 64.5             & 63.1             & 63.8             & 64.9               & 65.5              \\ \bottomrule
\end{tabular}
\caption{Results for the ablation of PANDA with Varied Training Data Quality. The baseline denotes the vanilla N-shot. TA denotes the accuracy of the training data (reflecting its quality). TA=82.4\% corresponds to labels predicted by the expert model, while TA=100\% represents ground truth labels.}
\label{tab:tweetTAwAblation}
\end{table*}

\subsection{ScienceWorld}
\subsubsection{Dataset Statistics}
To save time and cost while conducting extensive experiments on ScienceWorld, we adopted the same settings as~\citep{lin2023swiftsage}. Without loss of generality, we randomly select 2 tasks from each task type, resulting in a total of 20 tasks. For tasks with more than 50 variations in the training set, we randomly sampled 50 variations as the training set. Additionally, we randomly sampled up to 10 variations from the test set for performance evaluation. In Table~\ref{tab:SW-stat}, we provide detailed dataset statistics for ScienceWorld based on the aforementioned sampling approach.

\subsubsection{Prompt Template}
In this section, we present the prompt templates when implementing PANDA in ScienceWorld in Table~\ref{tab:SW-prompt-example1} and~\ref{tab:SW-prompt-example2}.
\subsection{TweetEval}
\subsubsection{Dataset Statistics}
Due to time and cost constraints, and without sacrificing generalization, we randomly sampled up to 1000 samples as the test data. As our goal with PANDA is to be a sample-efficient method, we use the same size for the training data. Detailed statistics for the dataset we use are presented in Table~\ref{tab:tweetDat}.

\subsubsection{Prompt Template}
In this section, we provide the prompt templates when implementing PANDA in TweetEval. We present prompt templates for both PANDA-Learning, PANDA-Inference and baselines in Fig.~\ref{tab:prompt-example1} to~\ref{tab:prompt-example4}.

\section{Discussion}
\subsection{Connection between PANDA and Conventional Knowledge Distillation}
\label{sec:panda-vs-kd}
PANDA and conventional knowledge distillation (KD) differ significantly in terms of implementation details. PANDA is a tuning-free approach, whereas conventional KD typically requires performing gradient descent to enable the student model to mimic the capabilities of the teacher model. Additionally, in conventional KD, the teacher model is usually superior to the student model, which is the opposite of the problem setting of this work. PANDA aims to enhance domain-specific capability of LLMs by leveraging the expertise of the expert model, while benefiting from the strong foundational capabilities of LLMs.

Despite these differences in technical details, we can demonstrate that PANDA and conventional KD share conceptual similarities at each stage. During the learning stage, conventional KD employs gradient descent to minimize the KL divergence between the output distributions of the student and teacher models:

\begin{equation*}
    \mathbf{\theta} = \arg\min_{\mathbf{\theta'}}\sum_{D}\mathrm{KL}(p_{\theta'},p_t),
\end{equation*}
where $\mathbf{\theta}$ and $\mathbf{\theta'}$ are the weights of the student model learned from the teacher model. $\mathrm{KL}(\cdot,\cdot)$ is the KL divergence. $p_{\theta'}$ and $p_t$ represent the output distribution of the student model and teacher model respectively. This minimization aims to reduce the knowledge gap between the student and teacher models, implicitly enhancing the capability of the student model. In contrast, PANDA achieves a similar goal but in a more direct manner. To comprehensively capture the expert knowledge, PANDA leverages insights derived from the expert, rather than relying solely on the raw responses, to gain a deep understanding of the expert knowledge. By incorporating insights from expert preferences, PANDA aims to enhance the knowledge comprehensiveness, going beyond capturing behavioral patterns of the expert alone.

PANDA also shares a similar essence with conventional KD during the inference stage. In conventional KD, the student model, with updated weights, is utilized for inference to achieve good performance. Similarly, leveraging the in-context learning capability of LLMs, PANDA incorporates relevant insights obtained from the expert preferences into the prompt context. This integration aims to enhance the capability of LLMs to effectively complete the given task. 

\begin{table*}[]
\centering\small
\begin{tabular}{lllrrr}
\toprule
\textbf{Task Type} & \textbf{Topic}          & \textbf{Name}                                       & \textbf{\#Vars: Train} & \textbf{\#Vars: Test} & \textbf{\# Actions} \\ \midrule
Task1-1   & Matter         & boil                                       & 14            & 9    & 273        \\
Task1-4   & Matter         & change-the-state-of-matter-of              & 14            & 9    & 233        \\
Task2-1   & Measurement    & use-thermometer                            & 50            & 10   & 227        \\
Task2-3   & Measurement    & measure-melting-point-unknown-substance    & 50            & 10   & 285        \\
Task3-1   & Electricity    & power-component (circuit)                  & 10            & 5    & 189        \\
Task3-4   & Electricity    & test-conductivity-of-unknown-substances    & 50            & 10   & 1,437       \\
Task4-1   & Classification & find-living-thing                          & 50            & 10   & 1,013       \\
Task4-2   & Classification & find-non-living-thing                      & 50            & 10   & 278        \\
Task5-1   & Biology        & grow-plant                                 & 50            & 10   & 2,728       \\
Task5-2   & Biology        & grow-fruit                                 & 50            & 10   & 2,137       \\
Task6-1   & Chemistry      & chemistry-mix                              & 16            & 8    & 136        \\
Task6-3   & Chemistry      & chemistry-mix-paint-tertiary-color         & 18            & 9    & 325        \\
Task7-1   & Biology        & lifespan-longest-lived                     & 50            & 10   & 205        \\
Task7-3   & Biology        & lifespan-longest-lived-then-shortest-lived & 50            & 10   & 228        \\
Task8-1   & Biology        & identify-life-stages-2 (plant)             & 4             & 4    & 109        \\
Task8-2   & Biology        & identify-life-stages-1 (animal)            & 6             & 5    & 37         \\
Task9-2   & Forces         & inclined-plane-friction-named-surfaces     & 50            & 10   & 395        \\
Task9-3   & Forces         & inclined-plane-friction-unnamed-surfaces   & 50            & 10   & 382        \\
Task10-1  & Biology        & mendelian-genetics-known-plant             & 50            & 10   & 1,377       \\
Task10-2  & Biology        & mendelian-genetics-unknown-plant           & 50            & 10   & 1,370       \\
\bottomrule
\end{tabular}

\caption{The statistics of ScienceWorld benchmark we use. We present the variations number of train-test splits for each task and the number of resulted actions in the training data.}
\label{tab:SW-stat}
\end{table*}

\begin{table*}[]
\centering\small
\begin{tabular}{l|rrr}
\toprule
\textbf{Task}            & \textbf{\# Train Data} & \textbf{\# Test Data} & \textbf{\# Class} \\ \midrule
\textbf{Emoji}           & 1,000                   & 1,000                  & 20                \\
\textbf{Emotion}         & 1,000                   & 1,000                  & 4                 \\
\textbf{Offensive}       & 860                    & 860                   & 2                 \\
\textbf{Sentiment}       & 1,000                   & 1,000                  & 3                 \\
\textbf{Hate}            & 1,000                   & 1,000                  & 2                 \\
\textbf{Irony}           & 784                    & 784                   & 2                 \\
\textbf{Stance-Abortion} & 280                    & 280                   & 3                 \\
\textbf{Stance-Atheism}  & 220                    & 220                   & 3                 \\
\textbf{Stance-Climate}  & 169                    & 169                   & 3                 \\
\textbf{Stance-Feminist} & 285                    & 285                   & 3                 \\
\textbf{Stance-Hillary}  & 295                    & 295                   & 3                 \\ \bottomrule
\end{tabular}
\caption{Dataset statistics of TweetEval we use.}
\label{tab:tweetDat}
\end{table*}

\begin{table*}[]
\centering
\begin{tabular}{l}
\toprule
\textbf{Prompt Template of PANDA-Inference for ScienceWorld}\\ 
\begin{tabular}[c]{@{}l@{}}
 \texttt{\{Init Prompt\}}\\
These are some insights that may be helpful for you to improve the success rate:\\
\texttt{\{Retrieved Insights\}}\\
 \texttt{\{Current Trajectory\}}\\
\end{tabular} \\ \bottomrule
\end{tabular}
\caption{Prompt template of PANDA-Inference for ScienceWorld.}
\label{tab:SW-prompt-example1}
\end{table*}

\begin{table*}[]
\centering
\begin{tabular}{l}
\toprule
\textbf{Prompt Template of PANDA-Learning for ScienceWorld}\\ 
\begin{tabular}[c]{@{}l@{}}
\texttt{\{Current Trajectory\}}\\
Now it is time to act again, the expert prefers \texttt{\{most preferred action\}} rather than\\ \texttt{\{second preferred action\}}. Please explain the reason why the expert holds on this preference.

\end{tabular} \\ \midrule
\textbf{Example of PANDA-Learning in ScienceWorld}\\ 
\begin{tabular}[c]{@{}l@{}}
\texttt{\#prompt here\#}\\ 
The expert's trial up to now is as follows:\\
Here is the task.\\
This room is called the kitchen. In it, you see:   the agent  a substance called air  a chair. On the chair is: \\
nothing.  a counter. On the counter is: a bowl (containing a red apple, a banana, an orange, a potato),\\
 a drawer. a cupboard. The cupboard door is closed.   a freezer. The freezer door is closed.   a fridge. \\
 The fridge door is closed.   a glass jar (containing a substance called sodium chloride)  a lighter\\
   a oven, which is turned off. The oven door is closed.   a painting  a sink, which is turned off. \\
   In the sink is: nothing.  a substance called soap  a stopwatch, which is deactivated.   a stove, \\
   which is turned off. On the stove is: nothing.  a table. On the table is: a glass cup\\
    (containing nothing).  a thermometer, currently reading a temperature of 10 degrees\\
    celsius You also see:  A door to the bathroom (that is open)  A door to the hallway (that is open)  A door\\
     to the outside (that is open)   Your task is to find the animal with the longest life span, then the \\
     shortest life span. First, focus on the animal with the longest life span.  Then, focus on the animal \\
     with the shortest life span. The animals are in the 'outside' location.\\
> go to outside\\
You move to the outside.; In your inventory, you see:  an orange ; This outside location is called the\\
 outside. Here you see:   the agent  a substance called air  an axe  a chameleon egg  a fire pit\\
  (containing nothing)  a fountain (containing a substance called water)  a giant tortoise egg\\
    the ground  a baby rabbit  a substance called wood You also see:  A door to the foundry (that is open)\\
      A door to the greenhouse (that is open)  A door to the kitchen (that is open) \\
> \\
Now it's time to act again, the expert prefers to focus on egg giant tortoise rather than to focus on \\chameleon. Please explain
 the reason why the expert holds on this preference.\\
Expert\_insight:\\
\texttt{\#response here\#}\\ 
INSIGHT: The expert prefers to focus on the giant tortoise egg rather than the chameleon egg because \\
the giant tortoise is known to have a significantly longer lifespan compared to the chameleon. \\
Giant tortoises are known to live for over 100 years, with some species even reaching ages of 150\\
 years or more. On the other hand, chameleons have a relatively shorter lifespan, typically ranging\\
  from 2 to 10 years depending on the species. Therefore, the expert believes that investigating\\
   the giant tortoise egg would be more likely to yield information about the animal with the longest lifespan.\\
\end{tabular} \\ \bottomrule
\end{tabular}
\caption{Prompt template and an example of PANDA-Learning for ScienceWorld.}
\label{tab:SW-prompt-example2}
\end{table*}

\begin{table*}[]
\centering
\begin{tabular}{l}
\toprule
\textbf{Prompt Template of PANDA-Inference for TweetEval}\\ 
\begin{tabular}[c]{@{}l@{}}
These are some insights that may be helpful for you to improve the success rate:\\
\texttt{\{Retrieved Insights\}}\\
\texttt{\{zero-shot/few-shot/cot prompt\}}\\
\end{tabular} \\ \bottomrule
\end{tabular}
\caption{Prompt template of PANDA-Inference for TweetEval.}
\label{tab:prompt-example1}
\end{table*}

\begin{table*}[]
\begin{tabular}{l}
\toprule
\textbf{Prompt Template of PANDA-Learning for TweetEval}\\ 
\begin{tabular}[c]{@{}l@{}}
\texttt{\#prompt here\#}\\ 
Output the \texttt{\{task name\}} of the given text.\\ 
Choose your answer from provided list and map your answer with following \texttt{\{mapping in task\}}\\ and return an integer as a result.\\
\texttt{\{Text\}}\\
Answer (\texttt{\{candidate answer\}}):\\
The expert prefers \texttt{\{most preferred answer\}} rather than \texttt{\{second preferred answer\}}.\\ Please explain the reason why the expert holds on this preference.\\
To determine the \texttt{\{task name\}} of a given text,\\
\texttt{\#response here\#}\\ 
INSIGHT:

\end{tabular} \\ \midrule
\textbf{Example of PANDA-Learning in TweetEval}\\ 
\begin{tabular}[c]{@{}l@{}}
\texttt{\#prompt here\#}\\ 
Output the sentiment of the given text. Choose your answer from provided list\\ and map your answer with following {negative: 0, neutral: 1, positive: 2}\\ and return an integer as a result.\\
Text: Today is Amazon Prime Day.  Today is the best day to buy.  July 15th only. \#savemoney\\
Answer (0 or 1 or 2):\\
The expert prefers positive(2) rather than neutral(1).\\ Please explain the reason why the expert holds on this preference.\\
To determine the sentiment of a given text,\\
\texttt{\#response here\#}\\ 
INSIGHT: To determine the sentiment of a given text,\\  the expert looks for words and phrases that convey a positive or negative tone.\\ In this text, the use of words like ``best day'' and ``save money'' convey a positive sentiment,\\ indicating that today is a good day to make purchases and take advantage of deals.\\ Therefore, the sentiment of the given text is positive (2).\\

\end{tabular} \\ \bottomrule
\end{tabular}
\caption{Prompt template and an example of PANDA-Learning.}
\label{tab:prompt-example2}
\end{table*}

\begin{table*}[]
\begin{tabular}{l}
\toprule
\textbf{Prompt Template of Zero-Shot for TweetEval}                               \\ 
\begin{tabular}[c]{@{}l@{}}
Output the \texttt{\{task name\}} of the given text.\\ 
Choose your answer from provided list and map your answer with following \texttt{\{mapping in task\}}\\ and return an integer as a result.\\
\texttt{\{Text\}}\\
Answer (\texttt{\{candidate answer\}}):\\
\end{tabular} \\ \midrule
\textbf{Prompt Template of Few-Shot for TweetEval}                                \\
\begin{tabular}[c]{@{}l@{}}
Output the sentiment of the given text. Choose your answer from provided list\\ and map your answer with following {negative: 0, neutral: 1, positive: 2}\\ and return an integer as a result.\\
Text: Dark Souls 3 April Launch Date Confirmed With New Trailer: Embrace the darkness.\\
Answer (0 or 1 or 2): 1\\
\\
Output the sentiment of the given text. Choose your answer from provided list\\ and map your answer with following {negative: 0, neutral: 1, positive: 2}\\ and return an integer as a result.\\
Text: ``National hot dog day, national tequila day, then national dance day... Sounds like a Friday night.''\\
Answer (0 or 1 or 2): 2\\
\\
Output the sentiment of the given text. Choose your answer from provided list\\ and map your answer with following {negative: 0, neutral: 1, positive: 2}\\ and return an integer as a result.\\
Text: When girls become bandwagon fans of the Packers because of Harry. \\Do y'all even know who Aaron Rodgers is?  Or what a 1st down is? \\
Answer (0 or 1 or 2): 0\\
\\
Output the \texttt{\{task name\}} of the given text.\\ 
Choose your answer from provided list and map your answer with following \texttt{\{mapping in task\}}\\ and return an integer as a result.\\
\texttt{\{Text\}}\\
Answer (\texttt{\{candidate answer\}}):\\ 
\end{tabular} \\ \bottomrule
\end{tabular}
\caption{Prompt template of zero-shot and few-shot for TweetEval.}
\label{tab:prompt-example3}
\end{table*}

\begin{table*}[]
\begin{tabular}{l}
\toprule
\textbf{Prompt Template of Zero-Shot CoT for TweetEval}                               \\ 
\begin{tabular}[c]{@{}l@{}}
Output the \texttt{\{task name\}} of the given text.\\    
Choose your answer from provided list and map your answer with following \texttt{\{mapping in task\}}\\ and return an integer as a result. Before output the answer, think it step by step firstly.\\
\texttt{\{Text\}}\\
Let's think step by step (end your answer with ``the answer is \#integer\#''):\\
\end{tabular} \\ \bottomrule
\end{tabular}
\caption{Prompt template of zero-shot Chain-of-Thought for TweetEval.}
\label{tab:prompt-example4-0}
\end{table*}

\begin{table*}[]
\begin{tabular}{l}
\toprule
\textbf{Prompt Template of Few-Shot CoT for TweetEval}                               \\ 
\begin{tabular}[c]{@{}l@{}}
Output the sentiment of the given text. Choose your answer from provided list\\ and map your answer with following {negative: 0, neutral: 1, positive: 2}\\ and return an integer as a result. Before output the answer, think it step by step firstly.\\
Text: Dark Souls 3 April Launch Date Confirmed With New Trailer: Embrace the darkness.\\
Let's think step by step (end your answer with ``the answer is \#integer\#''):\\
1. The text mentions the launch date being confirmed for Dark Souls 3, which is likely to be exciting\\news for fans of the game. This could indicate a positive sentiment.\\
2. The phrase ``Embrace the darkness'' could be interpreted as a call to action or a thematic element\\of the game, which may not necessarily indicate a negative sentiment.\\
3. Overall, the text seems to convey a sense of anticipation and excitement for the upcoming release.\\
Based on this analysis, the sentiment of the given text can be categorized as neutral. \\
Therefore, the answer is 1.\\
\\
Output the sentiment of the given text. Choose your answer from provided list\\ and map your answer with following {negative: 0, neutral: 1, positive: 2}\\ and return an integer as a result. Before output the answer, think it step by step firstly.\\
Text: ``National hot dog day, national tequila day, then national dance day... Sounds like a Friday night.''\\
Let's think step by step (end your answer with ``the answer is \#integer\#''):\\
Step 1: Identify key words and phrases\\
- ``National hot dog day''
- ``national tequila day''
- ``national dance day''
- ``Sounds like a Friday night''\\
Step 2: Analyze the tone and context\\
- The text seems to be expressing excitement or anticipation for the events mentioned.\\
- The mention of ``Friday night'' suggests a positive and fun atmosphere.\\
Step 3: Determine the sentiment
- The overall sentiment of the text is positive.\\
Step 4: Map the sentiment to the provided list
- Positive: 2\\
So, the answer is 2.\\
\\
Output the sentiment of the given text. Choose your answer from provided list\\ and map your answer with following {negative: 0, neutral: 1, positive: 2} and\\ return an integer as a result. Before output the answer, think it step by step firstly.\\
Text: When girls become bandwagon fans of the Packers because of Harry. \\  Do y'all even know who Aaron Rodgers is?  Or what a 1st down is? \\
Let's think step by step (end your answer with ``the answer is \#integer\#''):\\
Step 1: The text seems to be mocking girls who become fans of the Packers because of Harry,\\ implying that they may not actually know much about football.\\
Step 2: The tone of the text is somewhat negative and condescending.\\
Step 3: Based on the above analysis, the sentiment of the text is negative.\\
Step 4: Mapping the sentiment ``negative'' to the provided list, we get negative: 0.\\
Step 5: Therefore, the integer result for the sentiment of the text is 0. \\
So, the answer is 0.\\
\\
Output the \texttt{\{task name\}} of the given text.\\    
Choose your answer from provided list and map your answer with following \texttt{\{mapping in task\}}\\ and return an integer as a result. Before output the answer, think it step by step firstly.\\
\texttt{\{Text\}}\\
Let's think step by step (end your answer with ``the answer is \#integer\#''):\\
\end{tabular} \\ \bottomrule
\end{tabular}
\caption{Prompt template of few-shot Chain-of-Thought for TweetEval.}
\label{tab:prompt-example4}
\end{table*}

\end{document}